\documentclass[10pt,twocolumn,letterpaper]{article}

\usepackage[pagenumbers]{wacv}
\usepackage{times}
\usepackage{latexsym}
\usepackage[T1]{fontenc}
\usepackage[utf8]{inputenc}
\usepackage{microtype}
\usepackage{inconsolata}
\usepackage{graphicx}

\usepackage{amsmath}
\usepackage{amsfonts}
\usepackage{booktabs}
\usepackage{tcolorbox}
\usepackage{multicol}
\usepackage{algorithm}
\usepackage{algorithmic}
\usepackage{newfloat}
\usepackage{listings}
\usepackage{subcaption}
\usepackage{tabularx}
\usepackage{graphicx}
\usepackage{supertabular}
\usepackage{placeins}
\usepackage{float}
\usepackage{subcaption}
\usepackage{multirow}
\usepackage[font=small,labelfont=bf]{caption}
\usepackage{dblfloatfix}

\usepackage{booktabs}
\usepackage{tabularx}
\usepackage{array}
\usepackage{ragged2e}
\usepackage{amsmath}
\usepackage{enumitem}

\usepackage{booktabs}

\usepackage{xcolor}
\usepackage{tcolorbox}
\usepackage{longtable}
\usepackage{booktabs}
\usepackage{array}
\usepackage{ragged2e}

\definecolor{wacvblue}{rgb}{0.21,0.49,0.74}
\usepackage[pagebackref,breaklinks,colorlinks,allcolors=wacvblue]{hyperref}

\newcommand{\pearsonr}{\ensuremath{\mathit{r}}}
\tcbuselibrary{breakable}
\def\wacvPaperID{2816} % *** Enter the WACV Paper ID here
\def\confName{WACV}
\def\confYear{2027}

\title{Do VLMs See What Sensors Feel? A Scalable Expert-Guided Design for Wheelchair Accessibility Assessment from Street View\thanks{Accepted to the 2027 IEEE/CVF Winter Conference on Applications of Computer Vision (WACV 2027).}}

\author{
Dongdong Wang\textsuperscript{1} \quad
Alina Hagen\textsuperscript{2} \quad
Isabelle Gatmaitan\textsuperscript{1} \quad
Hao Zhou\textsuperscript{2} \\
Yiwen Dong\textsuperscript{3} \quad
Shabboo Valipoor\textsuperscript{1} \quad
Vivian W.H. Wong\textsuperscript{1}\thanks{Corresponding authors.} \quad
Lingyao Li\textsuperscript{4}\footnotemark[\value{footnote}]\\
{\small
\textsuperscript{1}University of Florida \quad
\textsuperscript{2}University of South Florida \quad
\textsuperscript{3}University of Illinois Urbana-Champaign
\quad
\textsuperscript{4}University of Arizona} \\
{\tt\small \{dongdongwang, igatmaitan,sh.valipoor,vivian.wong\}@ufl.edu} \\
{\tt\small \{alinahagen,haozhou1\}@usf.edu, yiwen@illinois.edu, lingyaoli@arizona.edu}
}

\begin{document}
\maketitle
\begin{abstract}

Assessing built-environment interaction, such as wheelchair accessibility, is challenging because real-world mobility is shaped by context-dependent and temporary barriers that are difficult to capture at scale. This paper examines whether vision-language models (VLMs) can identify potential accessibility barriers from Google Street View (GSV) imagery. We propose an expert-guided retrieval-augmented framework that combines GSV imagery, ADA-informed guidance, and expert-derived rubrics to evaluate multiple accessibility dimensions. We collect a campus-scale dataset at the University of Florida, linking 407 unique GSV locations with GPS-derived wheelchair dwell behavior as a proxy for potential mobility friction. Results show that VLM ratings are negatively correlated with dwell time and exhibit distributional similarity, with the expert-guided configuration strengthening the correlation from \(r=-0.40\) to \(r=-0.54\) for the strongest model. Visual cue analysis further shows that features such as curb ramps and crosswalks are associated with higher VLM accessibility scores, while correspondence remains limited for subtle surface conditions, transient obstructions, and viewpoint-dependent barriers. Overall, our findings demonstrate the potential of expert-guided VLMs for scalable preliminary accessibility screening and prioritizing locations for further investigation.

\end{abstract}    
\section{Introduction}
\label{sec:introduction}

% Wheelchair accessibility in urban systems is crucial for supporting independent movement, autonomy, and social participation among people with mobility disabilities. Although the Americans with Disabilities Act (ADA) and related accessibility standards have improved the baseline accessibility of built environments, such as buildings and public spaces, compliance with minimum standards does not always ensure that environments are comfortable, continuous, safe, or usable in everyday wheelchair navigation. Wheelchair users may still encounter uneven pavements, missing or poorly designed curb ramps, narrow or obstructed pathways, steep slopes, surface discontinuities, and other barriers that interrupt movement and limit access. Since these barriers are spatially distributed, context-dependent, and sometimes temporary, they are difficult to document through occasional or site-specific assessments alone. 

Wheelchair accessibility in urban systems is crucial for independent movement, autonomy, and social participation among people with mobility disabilities. Although the Americans with Disabilities Act (ADA) and related standards have improved baseline accessibility~\cite{doj2010ada}, minimum compliance does not always ensure environments are comfortable, continuous, safe, or usable in everyday navigation \cite{kapsalis2024disabled,prescott2021exploration}. Wheelchair users still encounter uneven pavements \cite{kapsalis2024disabled, duvall2013development}, poorly designed curb ramps \cite{kapsalis2024disabled,frost2010retrospective,wretstrand2010injuries,velho2016effect}, obstructed pathways, steep slopes \cite{d2019self}, surface discontinuities \cite{kapsalis2024disabled,bentzen2020effect}, and other barriers that interrupt movement. Because these barriers are spatially distributed, context-dependent, and sometimes temporary, they are difficult to document through occasional assessments \cite{froehlich2019grand}.

% Because these barriers are difficult to capture systematically, current assessments rely heavily on manual site visits. For instance, university accessibility offices often must cross-reference student schedules with physical audits of entire campus transit paths—a process that is both labor-intensive and unscalable. To overcome these limitations, we propose leveraging Google Street View (Street View) as a scalable visual data source for auditing outdoor urban environments. In this study, we investigate whether vision-language models (VLMs) can evaluate these environments by integrating visual cues, contextual reasoning, and domain-specific criteria (e.g., curb-ramp usability or surface conditions). Because accessibility judgments require applying design standards and understanding mobility nuances that may not be explicit in an image alone, we deploy VLMs as an expert-guided agent system to simulate accessibility assessors. An expert-guided approach is deployed to augment VLM reasoning with established accessibility \textbf{handbooks} and lived-experience-informed \textbf{rubrics}.

Because these barriers are difficult to capture systematically, current assessments rely heavily on manual site visits. For instance, university accessibility offices often cross-reference student schedules with physical audits of campus transit paths, a labor-intensive and unscalable process. %To address this limitation, we use Google Street View as a scalable visual source for auditing outdoor environments. 
To complement resource-intensive field assessment, we use Google Street View as a scalable visual source for screening visually observable built-environment conditions across outdoor environments. Prior studies suggest that VLMs can interpret Street View imagery for built-environment understanding, including urban perception and navigation-related reasoning~\cite{zhang2025urban, li2024georeasoner, ro2025well, wang2026cityvlm, schumann2024velma}. We therefore examine whether VLMs can evaluate accessibility by integrating visual cues, contextual reasoning, and domain-specific criteria. 
% Because accessibility judgments require design standards and mobility nuances not explicit in images alone, we deploy VLMs as an expert-guided agent system, augmenting their reasoning with accessibility handbooks and lived-experience-informed rubrics.

A further model-level gap is whether VLMs can interpret Street View imagery in ways that correspond to observed mobility behavior. This is central to accessibility assessment, where infrastructure appearance alone may not fully
reflect mobility friction during traversal. Although VLMs may identify problematic features in Street View imagery, it remains unclear whether visually inferred barriers align with \textbf{behavioral signals from wheelchair-based movement}, such as prolonged dwelling time. These gaps motivate two research questions:

\begin{itemize}[leftmargin=1em, itemsep=0pt, topsep=1pt, parsep=1pt, partopsep=1pt]
    \item \textbf{RQ1:} To what extent do VLM-derived accessibility assessments align with GPS-derived wheelchair dwell behavior, and how does expert-guided augmentation affect this correspondence?
    % To what extent do VLM-derived accessibility assessments align with sensor-derived mobility-friction signals, and how does expert-guided augmentation affect this alignment?
    %How well can VLMs identify wheelchair accessibility barriers from Street View imagery with and without expert-guided augmentation?
    \item \textbf{RQ2:} What geospatial patterns and visual cues explain correspondence and divergence between VLM-derived accessibility assessments and GPS-derived wheelchair dwell patterns?
    %What geospatial patterns and visual cues explain correspondence and divergence between VLM and wheelchair sensor patterns?
\end{itemize}

To answer these questions, we propose an expert-guided VLM system that assesses wheelchair accessibility from Street View imagery and compares its predictions with GPS-derived wheelchair dwell behavior on the University of Florida campus. Our contributions are: (i) a structured VLM pipeline for assessing potential accessibility barriers; (ii) an expert-guided augmentation strategy integrating accessibility standards and expert-informed criteria; and (iii) a campus-scale dataset linking Street View imagery with GPS-derived dwell observations to characterize correspondence and divergence. Together, these contributions support scalable preliminary accessibility screening.
% a campus-scale evaluation dataset linking Street View imagery with wheelchair-based mobility observations 
% %a campus-scale benchmark linking Street View imagery with wheelchair sensor traces 
% to identify where VLM-based assessments align with or diverge from sensor-derived mobility patterns. 
These contributions position our VLM pipeline as a scalable screening layer for accessibility assessment, helping urban planners and accessibility offices evaluate the built environment.
\section{Related Work}
\label{sec:literature}

% We frame related work as follows: prior work tells us \emph{what} to look for, but does not \emph{scale}; Existing VLM studies scale but are not \emph{grounded}. Our paper sits at the intersection.

\subsection{Existing accessibility assessment methods}

% \paragraph{Manual Audits.} While ADA standards have significantly improved the physical accessibility of built environments, they often fall short in fully accommodating the dynamic needs of users with mobility challenges. Field-audit approaches use ADA checklists to manually evaluate accessibility conditions in the built environment \citep{ada_checklist}. These methods provide explicit criteria for identifying barriers and remain central to facility audits, sidewalk assessments, and campus accessibility reviews, but they are often labor-intensive, episodic, primarily focused on minimum compliance rather than everyday wheelchair navigation, and not intended for end-user tools \citep{froehlich2019grand}.

\textbf{Manual Audits.} ADA standards have improved physical accessibility but often fail to address the dynamic needs of users with mobility challenges. Field-audit approaches use ADA checklists to manually evaluate accessibility~\cite{ada_checklist}, providing explicit criteria for identifying barriers and remaining central to facility audits, sidewalk assessments, and campus reviews. However, they are labor-intensive, episodic, focused on minimum compliance rather than everyday wheelchair navigation, and not designed for end-user tools~\cite{froehlich2019grand}.

\vspace{3px}\noindent\textbf{GIS/Crowdsourcing Platforms.} GIS-based inventories and routing platforms like AccessMap evaluate accessibility through connected infrastructure networks~\cite{li2018sidewalk, bolten2019accessmap}. To scale these efforts, crowdsourced systems like Project Sidewalk use Street View imagery and public participation~\cite{hara2013crowdsourcing}. However, both face data bottlenecks: GIS models depend on meticulously maintained infrastructure data, while crowdsourcing relies on static labels that lack coverage or direct evidence of wheelchair users' lived experiences.

% \vspace{2px}\noindent\textbf{GIS/Crowdsourcing Platforms.} Geographical Information System (GIS)-based inventories and routing platforms like AccessMap evaluate accessibility through connected infrastructure networks \citep{li2018sidewalk, bolten2019accessmap}. To scale these efforts, crowdsourced systems like Project Sidewalk utilize street-view imagery and public participation \citep{hara2013crowdsourcing}. However, both paradigms face significant data bottlenecks: GIS models depend on meticulously maintained infrastructure data, while crowdsourcing relies on static labels that often lack consistent coverage, temporal precision, and direct evidence of wheelchair users' actual lived experiences.

\vspace{3px}\noindent\textbf{Sensors.} Recent work explores wheelchair-mounted sensor systems to capture the embodied experience~\cite{gatmaitan2025}, such as inertial sensors/accelerometers~\cite{gendle2012wheelchair,pinnock2025inertial}, air quality sensors~\cite{park2022campus}, ego-centric cameras~\cite{kutbi2020usability}, and laser-enabled robotic arms~\cite{zhong2019assistive}. By logging continuous data during transit, these sensors identify dynamic, real-world barriers (e.g., cracked pavement, sudden maneuvers, inaccessible doors) that static visual audits miss. While this grounds accessibility assessment in actual user mobility, it requires physical traversal within the environment, reinforcing the need for complementary and scalable methods.

% \paragraph{Sensors.} To capture the embodied experience of navigating urban infrastructure, recent work has explored wheelchair-mounted sensor systems \cite{gatmaitan2025}. By logging continuous data during physical transit, sensors can identify dynamic, real-world barriers—such as cracked pavement, sudden maneuvers, or inaccessible doors—that static visual audits frequently miss. While this sensor-driven approach grounds accessibility assessment in actual user mobility, it inherently relies on active physical traversal of the environment. This reinforces the need for complementary, scalable visual screening methods capable of anticipating these barriers remotely.

\subsection{VLMs for built-environment study}

% With the emergence of large language models (LLMs), such as ChatGPT \cite{}, Gemini \cite{}, LLaMA \cite{touvron2023llama}, and Qwen \cite{bai2023qwen}, a new class of vision-language models (VLMs), including LLaVA \cite{liu2023visual}, Qwen-VL \cite{bai2023qwen}, and InternVL \cite{chen2024internvl}, has been developed to extend LLM capabilities by integrating visual perception with language reasoning, enabling advanced multimodal scene understanding and semantic environment analysis. Existing studies have demonstrated the effectiveness of VLMs in interpreting environmental context and supporting navigation-related reasoning \cite{li2024human, zhou2024navgpt}. 

% The success of VLM-based navigation systems further highlights the potential of multimodal reasoning for environmental awareness and spatial understanding, enabling scalable simulation and analysis in complex spatial environments. Recent studies further suggest that world-model-based frameworks can leverage visual understanding of GSV imagery to efficiently simulate large-scale environmental phenomena like earthquake magnitude \cite{li2025llms} with significantly reduced operational cost. These VLM-based approaches demonstrate strong potential for wheelchair accessibility assessment by enabling accurate interpretation of visual environments for scalable, explainable, and language-guided accessibility evaluation. Despite these advances, the use of VLMs for wheelchair accessibility assessment remains largely unexplored.

With the advance of large language models (LLMs), VLMs like LLaVA~\cite{liu2023visual}, Qwen-VL~\cite{bai2023qwen}, and InternVL~\cite{chen2024internvl} have been developed to extend LLM capabilities by integrating visual perception with language reasoning, enabling advanced multimodal scene understanding and semantic environment analysis~\cite{yang2024v, li2025pixels, feng2025urbanllava, howell2026multimodal}. In built-environment research, VLMs have been applied to Street View scene analysis, urban perception scoring, and navigation-related reasoning~\cite{li2024human, zhou2024navgpt, wang2026cityvlm}, demonstrating an ability to interpret environmental context from visual input. More recent work shows that VLMs can also serve as world-model simulators: Street View-based reasoning has been used to estimate large-scale environmental events such as earthquake shaking experience~\cite{li2025llms}, infer neighborhood health conditions~\cite{du2025advancing}, and assess perceived walkability and safety from street-level imagery~\cite{blevcic2024enhancing, tang2025street}. These results suggest that VLMs are well-positioned for scalable and explainable evaluation.

% Despite these advances, applying VLMs to wheelchair accessibility assessment remains largely unexplored. Existing applications can hallucinate, lack grounding in lived-experience knowledge, and are rarely evaluated alongside physical sensors, which is consequential when assessments inform mobility decisions. To address these gaps, we develop a scalable pipeline that grounds VLM reasoning over Street View imagery in design standards and 
% %validates it against wheelchair mobility patterns.
% evaluates its alignment with wheelchair mobility patterns.

Despite these advances, VLM-based wheelchair accessibility assessment remains underexplored. Existing models may hallucinate, lack expert-informed grounding, and are rarely evaluated against observed wheelchair mobility. We address these limitations through an expert-guided Street View assessment pipeline evaluated using GPS-derived dwell behavior.
\section{Data and Methods}
\label{sec:methods}

\begin{figure*}[!htbp]
    \centering
    \includegraphics[width=\linewidth]{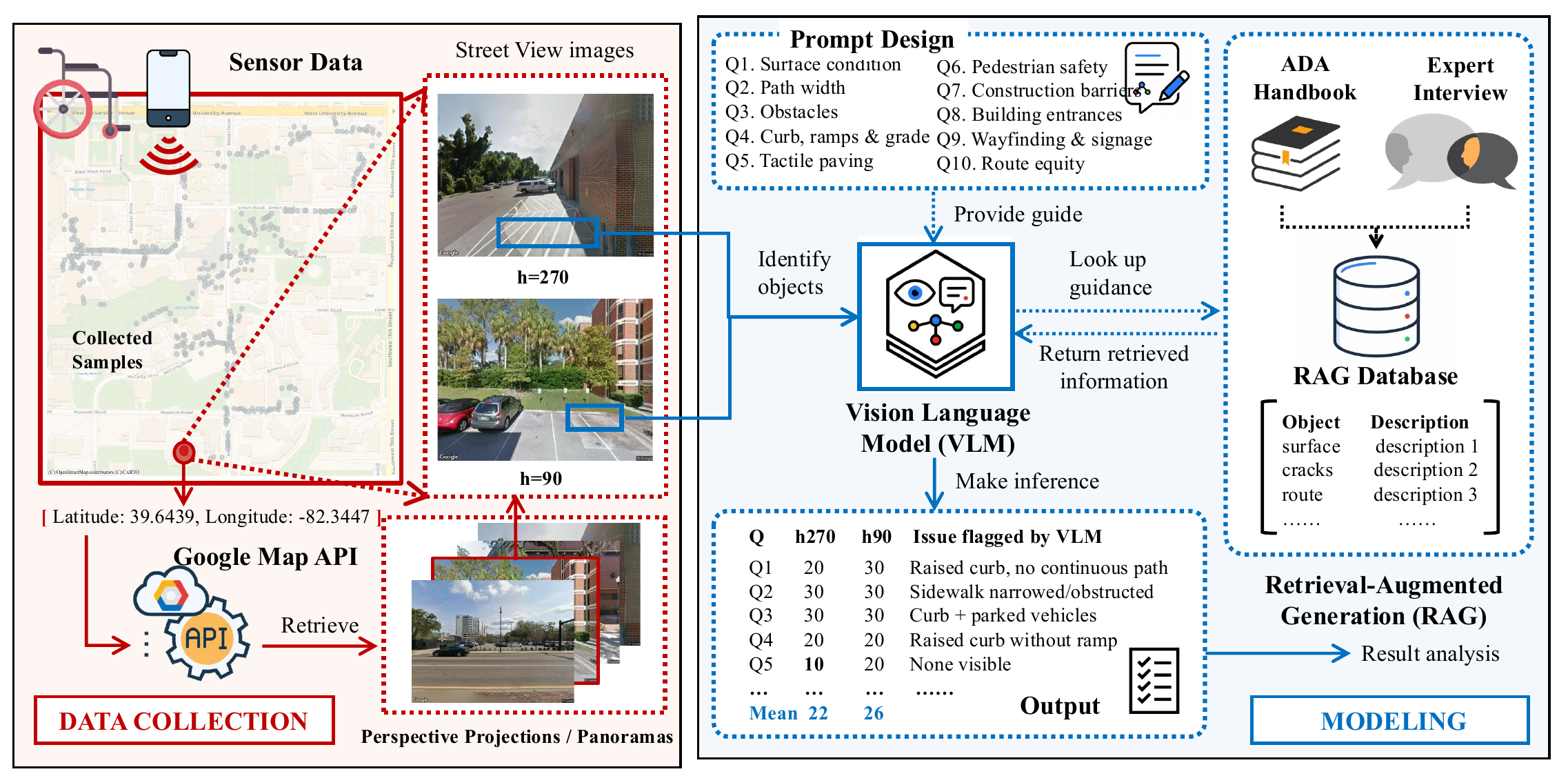}
    \caption{An illustration of research framework design. Sensor data refer to GPS observations.}
    \label{fig:research_framework}
    \vspace{-4mm}
\end{figure*}

Figure~\ref{fig:research_framework} illustrates our overall workflow. We first introduce the pilot study area in Section~\ref{sec:study_area}. We then describe wheelchair-mounted GPS data collection in Section~\ref{sec:sensor_data}, which is used to derive real-world mobility-friction patterns. Section~\ref{sec:gsv_image} explains how corresponding Street View images are retrieved along the same routes for scalable visual observations. Sensor data and Street View images capture complementary views of campus accessibility: observed wheelchair mobility and scalable visual context. Building on these paired data, Section~\ref{sec:vlm_pipeline} presents our expert-guided VLM pipeline for assessing potential accessibility barriers. Section~\ref{sec:vlm_output} describes how model outputs are designed to interpret accessibility scores and reasoning. Finally, Section~\ref{sec:evaluation} explains how we compare VLM-derived assessments with GPS-derived mobility-friction patterns to evaluate alignment.

\subsection{Study area}
\label{sec:study_area}

The study is conducted on the University of Florida (UF) campus in Gainesville, Florida, USA. UF provides a suitable setting for wheelchair accessibility assessment due to its heterogeneous built environment, including a mix of historic and modern infrastructure, varied sidewalk materials, curb ramps, and crosswalks. These varied route conditions frequently generate mobility barriers stemming from infrastructure design, surface quality, or temporary disruptions. Furthermore, as a major educational institution, the campus attracts a significant population of wheelchair users; it serves as a prime example of an environment that largely satisfies baseline ADA audits yet continues to present practical, everyday friction for wheelchair mobility.

To capture wheelchair-based mobility behavior, we collect sensor data along high-traffic routes connecting key campus destinations (e.g., classrooms, libraries, and the student union). By focusing on everyday travel rather than artificially constructed scenarios, we can meaningfully evaluate whether Street View-based assessments align with observed wheelchair mobility patterns.

\subsection{Wheelchair sensor data}
\label{sec:sensor_data}

We collect wheelchair-mounted GPS data across multiple campus trips using a smartphone attached to a manual wheelchair. Data collection was conducted under Institutional Review Board approval (Protocol \#ET00046077), with a researcher pushing an able-bodied participant in the manual wheelchair. The GPS streams record latitude and longitude during traversal and are used to reconstruct wheelchair trajectories and derive dwell behavior. 
%In this study, dwell time serves as a grounded mobility-friction signal for comparing Street View-based VLM accessibility ratings with observed wheelchair movement.
In this study, dwell time serves as an independently observed behavioral proxy for mobility friction, enabling evaluation of the alignment between Street View-based VLM accessibility ratings and observed wheelchair-based mobility behavior.

% We define dwell points as locations where the wheelchair remained within a localized area for an extended period rather than continuing along the route. These observations may reflect stationary periods or slow movement within a localized area. Longer dwell time may indicate environmental or infrastructural conditions that slow or interrupt navigation, such as inaccessible entrances, blocked paths, uneven terrain, construction, congestion, or route-finding around inaccessible features. However, prolonged dwell may also result from traffic signals, pedestrian interactions, or navigation decisions unrelated to accessibility barriers. Therefore, dwell time is treated as a proxy for potential mobility friction rather than ground-truth accessibility. The processed dataset contains 1,253 dwell point candidates and 7,201 non-dwell points, from which we retained a 10\% random sample of 720 non-dwell points for comparison. Details on dwell-point classification and contextual validation are provided in Appendices A and B of the supplementary material.

We define dwell points as locations where the wheelchair remained within a localized area rather than progressing along the route, potentially reflecting stationary periods or slow movement. Longer dwell may indicate obstructed paths, uneven terrain, construction, or other navigation difficulties. However, it may also result from traffic signals, pedestrian interactions, or navigation decisions unrelated to accessibility. We therefore treat dwell time as a proxy for potential mobility friction rather than ground-truth accessibility. The processed dataset contains 1,253 dwell candidates and 7,201 non-dwell points, from which we randomly sampled 720 non-dwell points for comparison. Classification and validation details are provided in Appendices A and B.

\subsection{Google Street View imagery}
\label{sec:gsv_image}

% Sampling strategy along the same routes (panorama interval, headings, time of capture), and how each image is linked to a sensor segment.

From the initial 8,454 collected locations, we perform a deduplication process to remove redundant Street View samples caused by the spatial resolution of Street View imagery. A 10-m distance threshold is applied to filter nearby duplicate locations. We further remove locations without valid Street View imagery, resulting in a final set of 407 unique spatial locations for image-based evaluation. For each location, we extract Street View imagery from eight panoramic viewing angles to provide comprehensive visual coverage of the surrounding. These locations serve as the spatial units for comparing VLM accessibility assessments with GPS-derived dwell observations.

% \begin{figure}[h]
% \includegraphics[width=0.95\columnwidth]{Figures/UF_sample_map.png}
% \caption{Data collection basemap showing the spatial distribution of wheelchair accessibility measurement sites used in the dataset. \DW{check the code for map} }
% \label{fig:sample_basemap}
% \end{figure}

\subsection{VLM pipeline design}
\label{sec:vlm_pipeline}

We develop a retrieval-augmented multimodal reasoning framework for wheelchair accessibility assessment using Street View imagery (Figure~\ref{fig:research_framework}). The pipeline first uses a VLM to identify accessibility-related environmental features (see Table S1 in Appendix F of the supplementary material), 
%(see Appendix~\ref{app:rag_lookup}), 
including curb ramps, sidewalks, crosswalks, surface conditions, and obstacles. These features are then used to retrieve relevant accessibility guidance from a compact knowledge base distilled from ADA standards, UF campus design guidelines, PROWAG, and expert-derived heuristics. The regulatory sources provide design-related criteria, while expert-derived heuristics offer additional contextual guidance for interpreting visible accessibility conditions. Retrieved information supports the VLM assessment but does not independently verify physical accessibility.

We evaluate three prompting configurations: (i) a base VLM with a generic accessibility prompt, (ii) handbook augmentation with retrieved regulatory guidance, and (iii) expert-opinion augmentation incorporating rubric-based knowledge from wheelchair users and accessibility professionals. To improve grounding reliability, the framework constrains reasoning to observable visual evidence while discouraging unsupported assumptions about non-visible infrastructure. The model evaluates sidewalk continuity, curb transitions, pedestrian safety, and obstacles, and outputs a structured accessibility score ranging from 10 (very poor) to 50 (excellent) with concise justification. Details of prompt and RAG design are presented in Appendices I and J of the supplementary material. %Appendices~\ref{app:rag_definitions} and \ref{app:prompt_design}.

\subsection{VLM output framework}
\label{sec:vlm_output}

We use a structured output schema to record per-image assessments of \emph{surface condition}, \emph{cross-slope}, \emph{obstructions}, \emph{curb-ramp presence and quality}, and \emph{width adequacy}, together with an overall accessibility score and a free-text justification. This schema makes VLM outputs comparable across images, aggregable across locations, and auditable during error analysis. The resulting scores represent image-based assessments of observable accessibility conditions rather than direct measurements of physical accessibility or wheelchair-user experience. %(see Appendix J of the supplementary material). %(see Appendix~\ref{app:prompt_design}).
We aggregate per-image assessments into location- and route-level accessibility scores for comparison with GPS-derived dwell observations, accounting for repeated Street View perspectives.

% We then aggregate per-image judgments into segment- and route-level accessibility scores. This aggregation can account for image-level confidence, repeated viewpoints, and spatial redundancy, allowing VLM-derived assessments to be compared with sensor-derived mobility segments. 

\begin{figure}[!ht]
    \centering
    \includegraphics[width=0.5\textwidth]{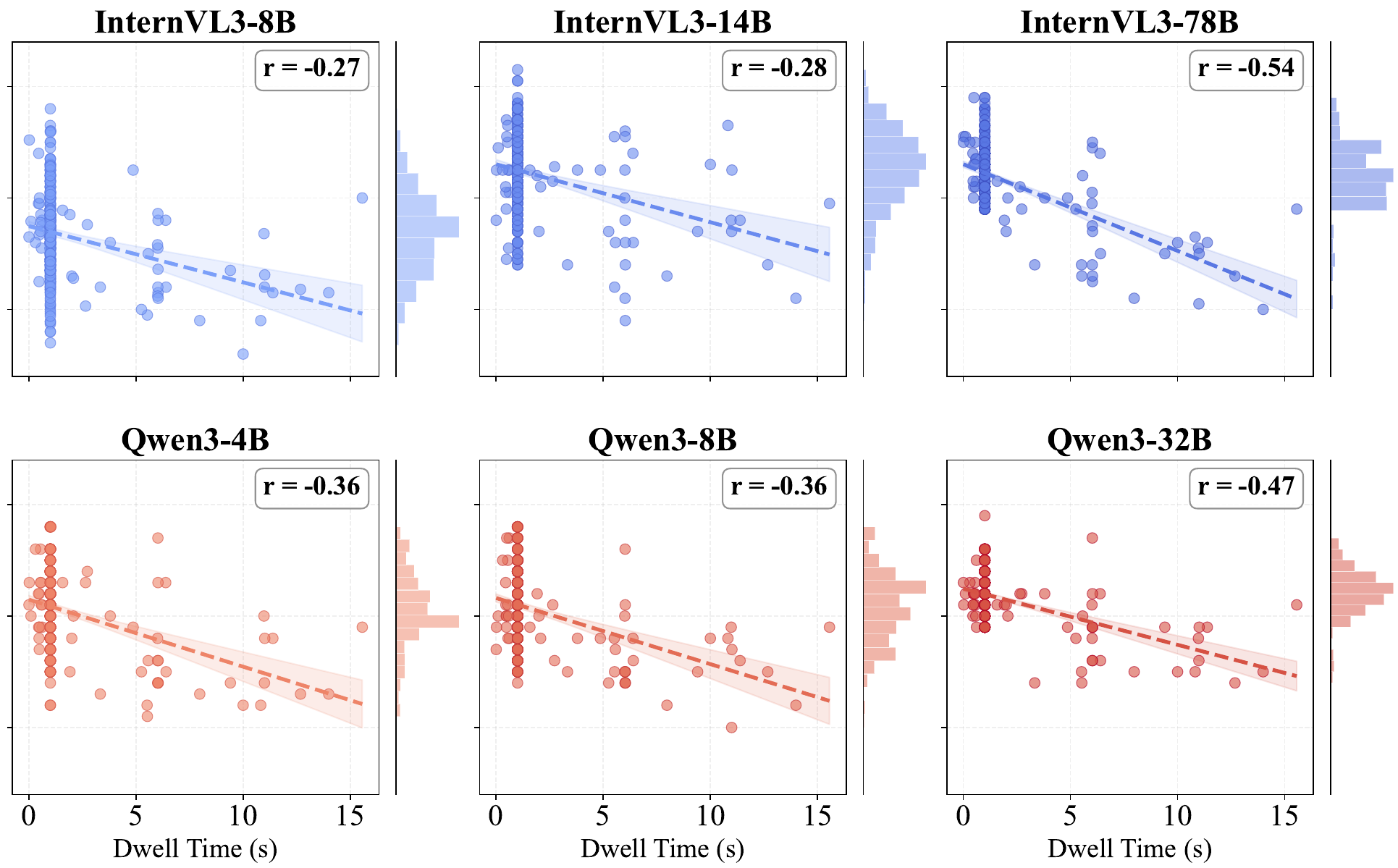}
    \vspace{-2mm}
    \caption{Correlation between wheelchair dwell time and VLM-based accessibility scores with RAG across 407 locations. }
    \label{fig:correlation_rag}
    \vspace{-3mm}
\end{figure}

\subsection{Evaluation}
\label{sec:evaluation}

% Agreement metrics against sensor ground truth (correlation, classification F1 for barrier/no-barrier), ablations over augmentation components, and spatial error analysis.

Alignment between sensor-derived mobility-friction patterns and VLM-based accessibility assessments is quantified using the Pearson correlation coefficient (\pearsonr{}):
% \begin{equation}
% r =
% \frac{
% \sum_{i=1}^{n}
% (x_i - \bar{x})
% (y_i - \bar{y})
% }{
% \sqrt{
% \sum_{i=1}^{n}(x_i - \bar{x})^2
% }
% \sqrt{
% \sum_{i=1}^{n}(y_i - \bar{y})^2
% }
% },
% \end{equation}
\begin{equation}
r =
\frac{
\sum_{i=1}^{n}(x_i-\bar{x})(y_i-\bar{y})
}{
\sqrt{\sum_{i=1}^{n}(x_i-\bar{x})^2}
\sqrt{\sum_{i=1}^{n}(y_i-\bar{y})^2}
}
\end{equation}
where \(x_i\) denotes the GPS-derived indicator of wheelchair mobility friction, which is dwelling time in seconds (as stated in Section \ref{sec:sensor_data}), \(y_i\) represents the corresponding VLM-generated accessibility score, \(\bar{x}\) and \(\bar{y}\) are the sample means, and \(n\) is the total number of observations. Because longer dwell times 
%reflect greater mobility friction
can indicate greater mobility friction, a strong negative correlation (\pearsonr{} approaching $-1$) indicates greater correspondence 
%high agreement 
between the observed mobility behavior and the VLM's multimodal accessibility reasoning. However, this correlation measures association with dwell behavior rather than agreement with independently verified accessibility barriers, as prolonged dwell may also reflect non-accessibility-related factors. We compare \pearsonr{} across different augmentation and multimodal input configurations to evaluate which setup yields accessibility assessments that align most closely with observed wheelchair-based mobility patterns.
%align most with patterns observed in real-world wheelchair navigation.

We also assess distributional correspondence between normalized dwell time and VLM-generated accessibility scores using the Earth Mover’s Distance (EMD), also known as the Wasserstein distance:
\begin{equation}
\begin{aligned}
\mathrm{EMD}(X,Y)
&=
W_1(X,Y) \\
&=
\inf_{\gamma \in \Pi(X,Y)}
\int_{\mathbb{R}\times\mathbb{R}}
|x-y| \, d\gamma(x,y),
\end{aligned}
\end{equation}
where \(X\) and \(Y\) denote the normalized distributions of dwell time and VLM accessibility scores, respectively, and \(\Pi(X,Y)\) represents the set of all joint distributions with marginals \(X\) and \(Y\). Both dwell time and VLM accessibility scores are independently min–max normalized to \([0,1]\). To account for their opposite semantic directions, normalized dwell time is inverted before computing EMD. Lower EMD values indicate greater similarity between the resulting marginal distributions, but do not necessarily imply better location-level correspondence or more accurate identification of accessibility barriers. Given the imbalanced and long-tailed distribution of mobility sensing data, we report both overall EMD and dwell-time quantile-specific EMD, with particular attention to Q4, which contains the longest dwell times.

We conduct two complementary visual cue association analyses to examine semantic objects associated with VLM--sensor alignment and accessibility scoring patterns. For each object \(o\), we compute rank differences between top- and bottom-ranked 10\% regions under different criteria:

\vspace{-1mm}
\begin{align}
AMI_o &=
\operatorname{rank}_{\mathrm{align}}(o)
-
\operatorname{rank}_{\mathrm{misalign}}(o), \\
AII_o &=
\operatorname{rank}_{\mathrm{access}}(o)
-
\operatorname{rank}_{\mathrm{inaccess}}(o).
\end{align}
where $\mathrm{rank}_{align}(o)$ and $\mathrm{rank}_{misalign}(o)$
denote object-frequency ranks in well- and poorly-aligned regions,
while $\mathrm{rank}_{access}(o)$ and $\mathrm{rank}_{inaccess}(o)$
denote ranks among locations receiving high and low VLM accessibility
scores, respectively. Smaller ranks indicate higher object frequency;
thus, negative values indicate stronger association with aligned
locations or locations receiving higher VLM accessibility scores.

\section{Results}
\label{sec:results}

\subsection{Pipeline performance assessment}
\label{sec:quant_pipeline}

\noindent\textbf{Quantitative Analysis.} Figure~\ref{fig:correlation_rag} presents Pearson correlations between dwell time and VLM accessibility scores, with trip-level results in Table S2. %Table~\ref{tab:trip_correlations}.
Consistent negative correlations are observed across the InternVL and Qwen model families. We also evaluate GPT-4.1 as a representative proprietary multimodal model (Table \ref{tab:rag_vs_base}). GPT-4.1 achieves \(r=-0.46\), Overall EMD \(=0.447\), and Q4 EMD \(=0.288\), comparable to the strongest open-source models and suggesting that the observed pattern extends beyond the two evaluated open-source model families. This indicates that locations associated with prolonged wheelchair dwell  times receive correspondingly lower VLM accessibility scores. These findings suggest that VLM scores capture accessibility-related characteristics of the built environment that partially align with GPS-derived dwell patterns.
%These findings demonstrate the capability of VLMs to capture accessibility-related characteristics of the built environment from Street View imagery, aligning meaningfully with real-world mobility friction behavior.

Further, larger models generally demonstrate stronger alignment with GPS-derived mobility friction signals, with InternVL3-78B achieving the strongest correlation (\(\pearsonr=-0.54\)) followed by Qwen3-32B (\(\pearsonr=-0.47\)). In contrast, smaller variants such as InternVL3-2B and Qwen3-2B exhibit substantially weaker correlations (\(\pearsonr=-0.13\) and \(\pearsonr=-0.10\), respectively), indicating limited alignment with observed mobility friction patterns. These results suggest that larger model variants are generally more sensitive to potential accessibility barriers associated with GPS-derived dwell patterns.

% The results suggest that greater model capacity improves a VLM's sensitivity to potential accessibility barriers associated with 
% %reflected by 
% GPS-derived dwell patterns.

% How well the full pipeline scores accessibility, with ablations isolating the contribution of (a) handbook retrieval and (b) expert rubric. Headline finding: both augmentations help, ideally in complementary ways (handbook $\rightarrow$ regulatory features; expert rubric $\rightarrow$ lived-experience features the handbook misses).

\begin{table}[!htbp]
\centering
\caption{RAG versus non-RAG prompting performance comparison using \pearsonr{} and EMD.}
\label{tab:rag_vs_base}
\small
\setlength{\tabcolsep}{2pt}
\renewcommand{\arraystretch}{1.05}

\begin{tabular}{llcccc}
\toprule
\textbf{Model Variant} & \textbf{Setting} & \textbf{\pearsonr{}$\downarrow$} & \textbf{Overall EMD$\downarrow$} & \textbf{Q4 EMD$\downarrow$} \\
\midrule

InternVL3-78B & Non-RAG & -0.40 & 0.592 & 0.421 \\
InternVL3-78B & RAG     & -0.54 & 0.530 & 0.337 \\
\hline
Qwen3-VL-32B  & Non-RAG & -0.40 & 0.517 & 0.345 \\
Qwen3-VL-32B  & RAG     & -0.47 & 0.420 & 0.246 \\
\hline
GPT-4.1	& Non-RAG	& -0.38	& 0.533 & 0.376 \\
GPT-4.1	& RAG	& -0.46	& 0.447 & 0.288 \\

\bottomrule
\end{tabular}

\vspace{-3mm}
\end{table}

\begin{figure*}[h]
    \centering
         \includegraphics[width=0.93\linewidth]{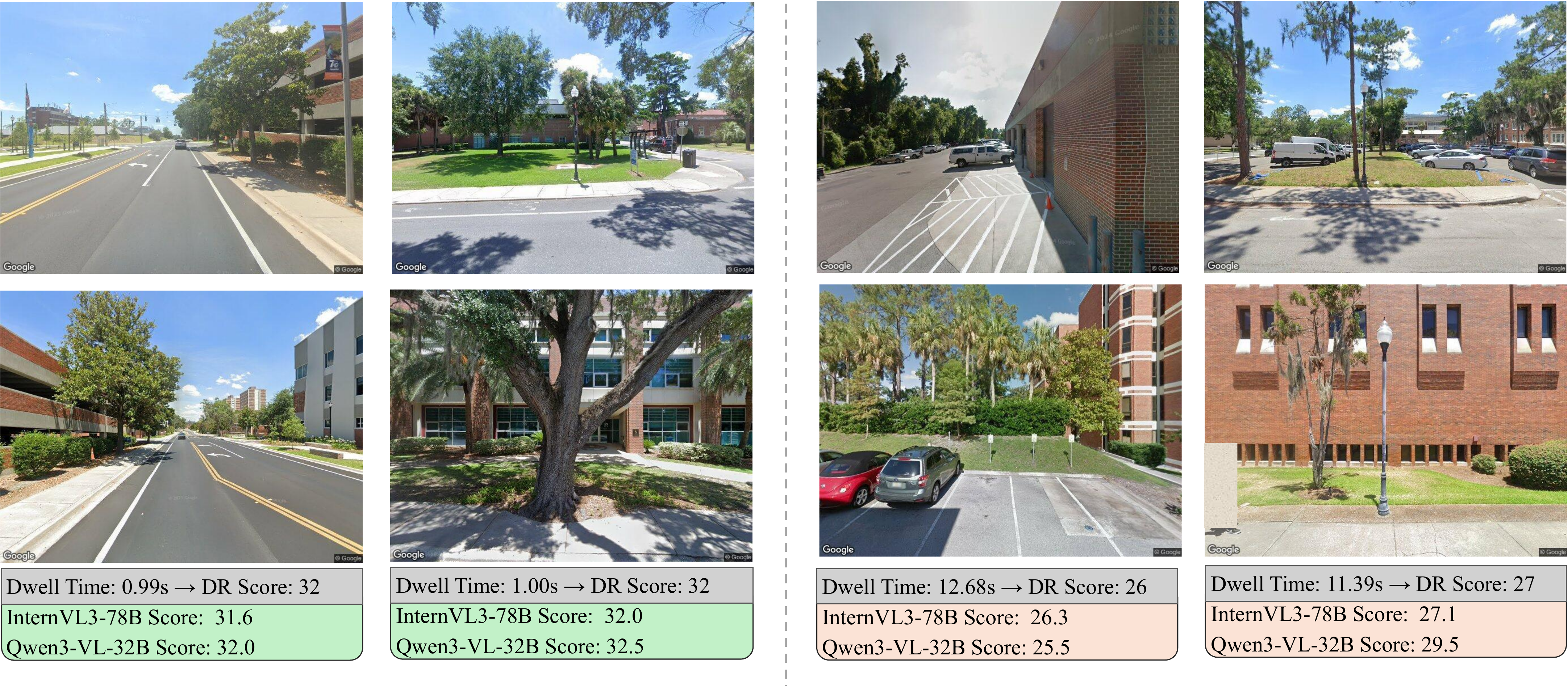}
    \vspace{-2mm}
    \caption{Representative success cases showing strong correspondence between dwell behavior and VLM-derived accessibility scores. Columns 1–2 show low-dwell locations with relatively high VLM scores, while Columns 3–4 show high-dwell locations with relatively low VLM scores. Dwell-derived reference (DR) scores are estimated from the fitted dwell–score relationship for visualization only and do not represent ground-truth accessibility labels.}
    % \caption{Success cases illustrating strong alignment between dwell time and VLM-derived accessibility scores. Columns 1–2 demonstrate accessible locations, while Columns 3–4 illustrate inaccessible locations. Dwell-derived reference (DR) scores are estimated using the fitted dwell–score correlation function.} 
    % Ground-truth (GT) scores are estimated using the fitted correlation function.}
    \vspace{-2mm}
    \label{fig:visual_success_cases}
\end{figure*}

\begin{figure*}[h]
    \centering
         \includegraphics[width=0.93\linewidth]{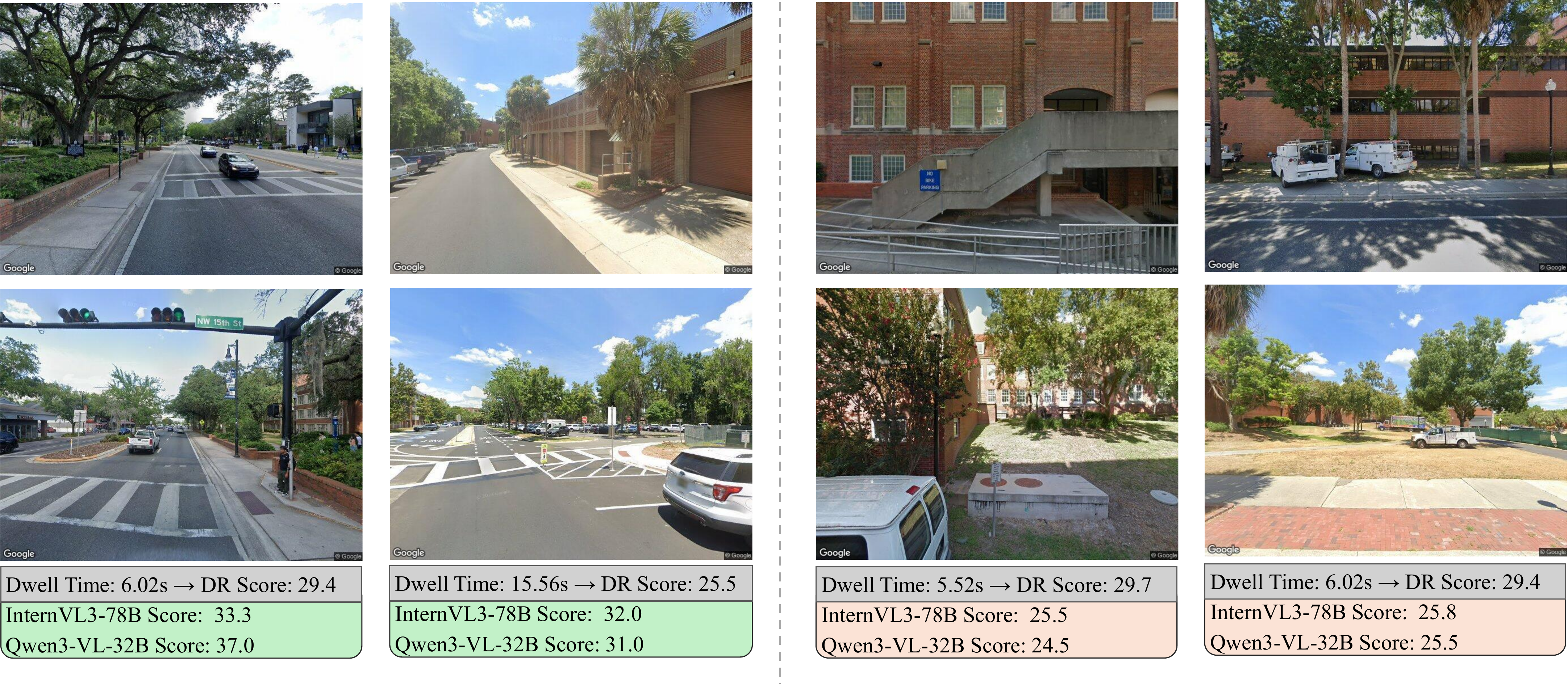}
    \vspace{-2mm}
    \caption{Representative cases illustrating misalignment between dwell behavior and VLM-derived accessibility scores. Columns 1-2 show cases where VLM scores are higher than expected from the fitted dwell–score relationship, while Columns 3-4 show the opposite pattern. DR scores are used only to visualize the fitted relationship and do not represent ground-truth accessibility labels.}
    %\caption{Failure cases illustrating weak alignment between dwell time and VLM-derived accessibility scores. Columns 1–2 demonstrate accessibility underestimation, while Columns 3–4 illustrate accessibility overestimation. Dwell-derived reference (DR) scores are estimated using the fitted dwell–score correlation function.} % 
    \vspace{-2mm}
    \label{fig:visual_failure_cases}
\end{figure*}

\vspace{3px}\noindent\textbf{Qualitative Analysis.}
Figures~\ref{fig:visual_success_cases} and~\ref{fig:visual_failure_cases} provide representative examples illustrating success and failure cases, while Figure S2 in Appendix D of the supplementary material 
%Figure~\ref{fig:visual_cue} in Appendix 
presents additional qualitative examples illustrating the relationship between mobility sensing behavior and multimodal accessibility reasoning. The results demonstrate that both InternVL3-78B and Qwen3-VL-32B are sensitive to contextual pedestrian accessibility cues and partially reflect observed mobility behavior. In low dwell-time examples (\(1.00s\)), both models assign relatively higher accessibility scores (\(30\)–\(33\)), emphasizing smooth sidewalks, wider pathways, curb ramps, crosswalks, and fewer severe obstructions. In contrast, high dwell-time examples (\(12.68s\)) consistently receive lower scores (\(22\)–\(27\)), where both models identify fragmented pedestrian environments including raised curbs, missing curb ramps, constrained pathways, temporary barriers, and obstructed movement.

Figure~\ref{fig:visual_success_cases} shows that the models effectively recognize accessibility-supportive pedestrian infrastructure and produce accessibility scores that closely align with GPS-derived mobility friction patterns. However, the failure cases in Figure~\ref{fig:visual_failure_cases} reveal limitations in complex environments involving ambiguous pedestrian continuity, viewpoint-dependent cues, transient obstructions, or subtle surface degradation, where models occasionally deviate from the observed mobility-friction patterns. This shows that VLMs capture key accessibility-related environmental features while remaining sensitive to subtle pedestrian conditions and visual complexity.

\subsection{Ablation study}
\label{sec:ablation}

\noindent\textbf{RAG versus Non-RAG Reasoning.}
We first examine whether RAG improves alignment between VLM-based accessibility ratings and observed wheelchair mobility behavior. Compared with base prompting using generic accessibility instructions (Figure S3 in Appendix E of the supplementary material),
%(Appendix~\ref{app:baseline_correlation})
the expert-guided RAG configuration consistently yields stronger negative correlations and lower EMD across all model variants. As shown in Table~\ref{tab:rag_vs_base}, RAG improves \pearsonr{} from $-0.40$ to $-0.54$ for InternVL3-78B and from $-0.40$ to $-0.47$ for Qwen3-VL-32B, indicating stronger alignment between accessibility ratings and GPS-derived dwell time. These improvements reflect the combined augmented configuration and do not isolate the individual contributions of retrieval and domain-specific guidance. RAG also reduces overall EMD from 0.592 to 0.530 for InternVL3-78B, from 0.517 to 0.420 for Qwen3-VL-32B, and from 0.533 to 0.447 for GPT-4.1, while decreasing Q4 EMD from 0.421 to 0.337, from 0.345 to 0.246, and from 0.376 to 0.288, respectively. The higher overall EMD values are partly influenced by the imbalanced score distribution, with most samples concentrated in the middle range of VLM accessibility scores. In contrast, Q4 EMD focuses on locations with higher dwell times, emphasizing the high-friction portion of the mobility distribution. For the stronger models, Q4 EMD values decrease to approximately 0.25-0.35. Overall, these results suggest that grounding multimodal reasoning with retrieved ADA handbook knowledge and expert-derived guidance improves both correlation-based and distributional alignment between VLM accessibility assessment and GPS-derived wheelchair mobility friction.

\begin{figure}[!ht]
    \centering
    \vspace{-1mm}
    \includegraphics[width=\columnwidth]
    {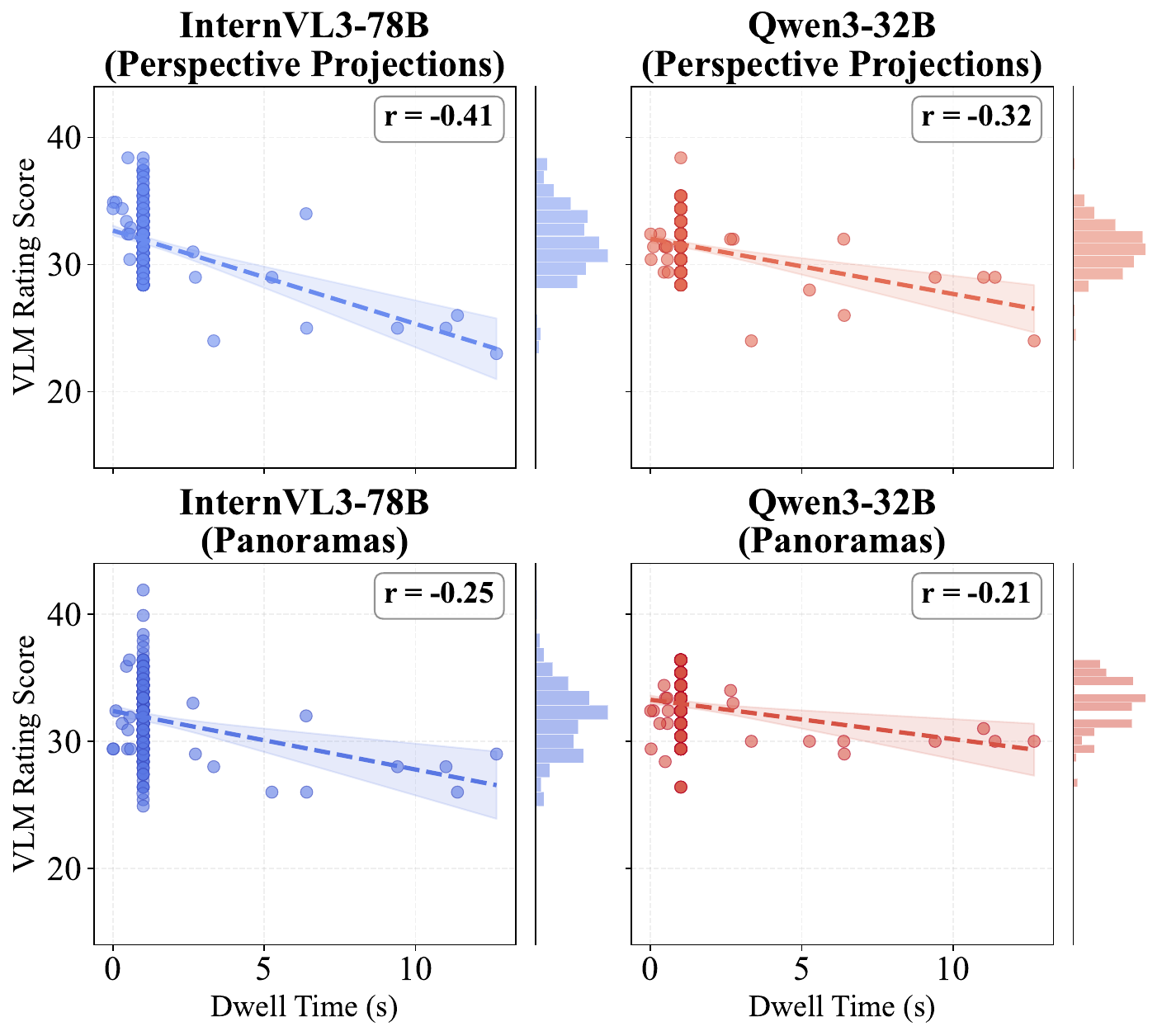}
    \caption{Correlation coefficients of VLM-generated ratings using 90-270 perspective projection (above) and panorama (below) Street View inputs across 246 sampled locations.}
    \label{fig:correlation_streetview_angle}
    \vspace{-4mm}
\end{figure}

\vspace{3px}\noindent\textbf{Street View Perspective Projections versus Panoramas.}
We next examine how Street View input formats affect accessibility reasoning by comparing perspective projections (paired \(h{=}90^{\circ}\) and \(h{=}270^{\circ}\) perspectives) with full panoramas (Figure~\ref{fig:correlation_streetview_angle}). Across both models, perspective projection inputs yield consistently stronger alignment with GPS-derived dwell time than panorama inputs. For example, InternVL3-78B achieves \(r = -0.41\) with perspective projections compared to \(r = -0.25\) with panorama inputs, and Qwen3-VL-32B shows a similar pattern (\(r = -0.32\) vs. \(r = -0.21\)).

\begin{figure*}[!ht]
    \centering
    \includegraphics[width=\textwidth]{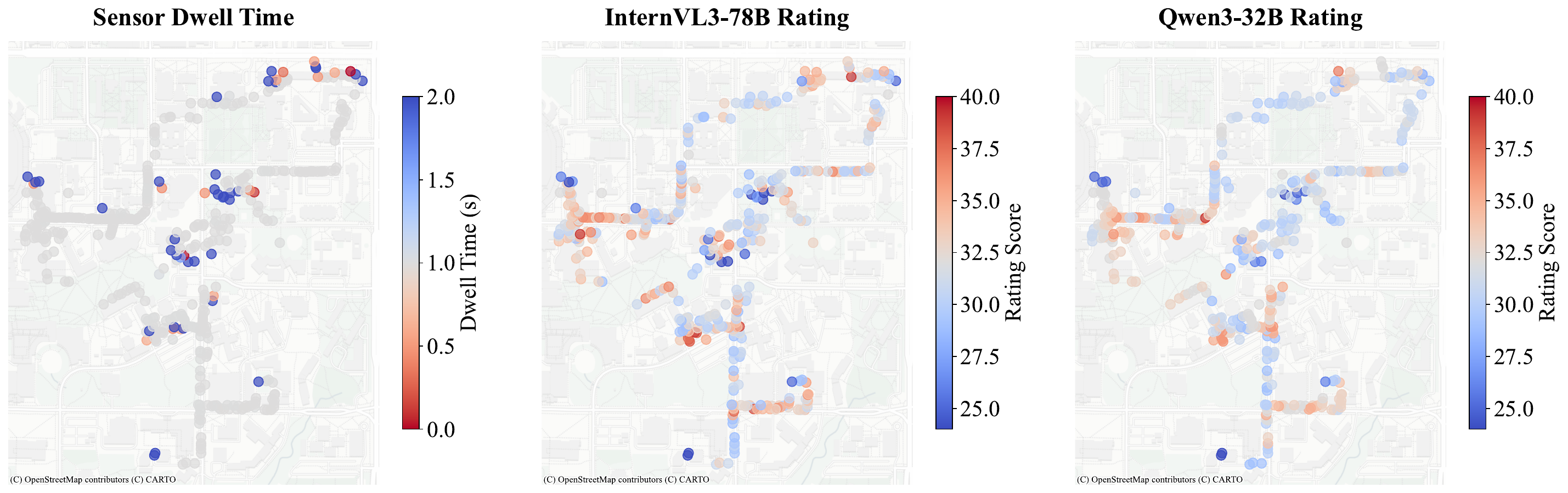}
    \caption{Spatial comparison of GPS-derived wheelchair dwell time and VLM-derived accessibility scores.}
    % \caption{Spatial distribution comparison between measured dwell time and VLM accessibility scores.}
    \label{fig:spatial_distribution}
    \vspace{-4mm}
\end{figure*}

We hypothesize that this performance gap stems from both geometric and data limitations. Physically, the distortion of spherical projections can obscure fine-grained visual cues (e.g., surface texture, curb transitions, boundaries of obstructions), and panoramic views, which aggregate omnidirectional information, dilute attention compared to wide-angle configurations in perspective projection images that more closely align with the forward direction of travel. Methodologically, VLMs are likely under-trained on panoramas compared to standard perspective-projection imagery

% One hypothesis is that explains the performance gap is that distortion and geometric compression introduced by spherical projection can obscure fine-grained visual cues, such as surface texture, curb transitions, and obstruction boundaries, which are more clearly preserved in wide-angle imagery. Additionally, the dual-perspective wide-angle configuration provides directional context that aligns more naturally with how wheelchair users encounter accessibility features along a path, whereas panorama views aggregate omnidirectional information that may dilute attention on the immediate pedestrian route. An alternate hypothesis is that VLMs are not trained with as much Street View panorama images compared to the more common perspective projection images. 

\begin{table}[h]
\centering
\caption{Single- (SQ) versus multi- (MQ) question prompting performance measured by \pearsonr{} and EMD. }
\label{tab:single_vs_multi_question}
\small
\setlength{\tabcolsep}{2pt}
\renewcommand{\arraystretch}{1.05}

\begin{tabular}{llcccc}
\toprule
\textbf{Model Variant} & \textbf{Setting} & \textbf{\pearsonr{} $\downarrow$} & \textbf{Overall EMD $\downarrow$} & \textbf{Q4 EMD $\downarrow$} \\
\midrule

InternVL3-78B & SQ & -0.213 & 0.584 & 0.481 \\
InternVL3-78B & MQ   & -0.540 & 0.530 & 0.337 \\
\hline
Qwen3-32B     & SQ & -0.268 & 0.860 & 0.779 \\
Qwen3-32B     & MQ  & -0.470 & 0.420 & 0.246 \\

\bottomrule
\end{tabular}

\vspace{-3mm}
\end{table}

\vspace{3px}\noindent\textbf{Multi-Question versus Single-Question.} 
Table~\ref{tab:single_vs_multi_question} shows that multi-question prompting substantially improves alignment between VLM accessibility ratings and mobility-derived behavioral signals compared to single-question prompting. InternVL3-78B improves from \(\pearsonr=-0.213\) under single-question prompting to \(\pearsonr=-0.54\) with ten-question reasoning, while Qwen3-32B improves from \(\pearsonr=-0.268\) to \(\pearsonr=-0.470\), with similar patterns observed across the EMD metric. The results suggest that decomposing wheelchair accessibility assessment into structured sub-questions enables more reliable multimodal reasoning and improves sensitivity to accessibility-related environmental cues.

\vspace{3pt}
\noindent\textbf{High-Friction Location Retrieval.}
We further evaluate the VLM scores for identifying locations with high dwell time. For retrieval evaluation, high-friction locations are operationally defined by GPS-derived dwell duration using the 75th- and 90th-percentile dwell-time thresholds, and we report AUROC, Precision@20, and Recall@20. Here, $20$ corresponds to approximately 5\% of the 407 study locations. These metrics evaluate the identification of high-dwell locations rather than independently verified physical accessibility barriers. As shown in Table~\ref{tab:high_friction_retrieval}, at the 90th-percentile threshold, InternVL3-78B achieves an AUROC of 0.835, 100.0\% Precision@20, and 48.8\% Recall@20. GPT-4.1 achieves comparable performance, with an AUROC of 0.842, 95.0\% Precision@20, and 46.3\% Recall@20. These results show that VLM scores can be used to prioritize high-friction locations for subsequent accessibility review.

\begin{table}[h]
\centering
\footnotesize
\setlength{\tabcolsep}{3pt}
\caption{High-friction location retrieval performance using dwell-time percentile thresholds. For reference, random selection yields an expected Precision@20 of approximately 10$\%$ at the 90th-percentile threshold. }
\label{tab:high_friction_retrieval}
\begin{tabular}{lcccc}
\toprule
Model & Threshold & AUROC $\uparrow$ & Precision@20 $\uparrow$ & Recall@20 $\uparrow$ \\
\midrule
InternVL3-78B & $>75$th & 0.726 & 100.0\% & 19.6\% \\
GPT-4.1       & $>75$th & 0.650          & 95.0\%          & 18.6\% \\
\hline
InternVL3-78B & $>90$th & 0.835          & 100.0\% & 48.8\% \\
GPT-4.1       & $>90$th & 0.842 & 95.0\%          & 46.3\% \\
\bottomrule
\end{tabular}
\end{table}

% \begin{figure*}[t]
%     \centering
%     \includegraphics[width=\textwidth]{Figures/outlier_inlier_separate_bar_charts.png}
%     \caption{Detected object frequencies in outlier and inlier locations identified from residual-distance analysis between accessibility scores and sensor-derived mobility behavior. Outlier locations correspond to the top 20\% largest residual distances from the regression line, while inlier locations correspond to the bottom 10\% closest-fitting locations.}
%     \label{fig:word_frequency}
%     \vspace{-3mm}
% \end{figure*}

\subsection{Geospatial and visual cue analysis}
\label{sec:geospatial_analysis}

% Quantitative agreement at segment and route levels; where the LLM and sensors agree (typically obvious barriers and clearly compliant infrastructure) and where they diverge (often subtle surface conditions, transient obstructions, or features only visible from a wheelchair viewpoint). A campus-level map makes the spatial story concrete.

\vspace{3px}\noindent\textbf{Geospatial Pattern.} Quantitative alignment is evaluated at both measurement and trip levels through correlations between mobility sensing signals and VLM accessibility scores. Figure~\ref{fig:spatial_distribution} further visualizes the campus-scale spatial relationship between GPS-derived dwell time and VLM accessibility score predictions. The models generally align with sensor observations in regions containing obvious accessibility barriers or accessibility-supportive pedestrian infrastructure.
%clearly compliant pedestrian infrastructure. %, whereas larger discrepancies frequently emerge in environments involving subtle surface degradation, transient obstructions, or accessibility cues that are difficult to perceive from standard Street View perspectives. 
The spatial distributions suggest that VLMs capture potential accessibility barrier location patterns. %while remaining sensitive to complex real-world pedestrian conditions.

% \subsection{Visual Cues Analysis}
\label{sec:visual_cues}

%We further propose an AMI to analyze semantic object associations between well-fitted and poorly-fitted accessibility predictions. Specifically, we first partition locations into low and high residual-distance groups based on the disagreement between mobility sensing signals and LLM accessibility scores. Detected objects are then ranked according to their occurrence frequency within each group. The proposed index computes the signed rank difference between aligned and misaligned regions, where positive values indicate stronger association with semantic misalignment, negative values indicate stronger association with semantic alignment, and values near zero represent neutral contribution. This analysis provides an interpretable semantic perspective on which urban objects potentially contribute to discrepancies between human mobility behavior and multimodal accessibility reasoning.

\begin{figure}[h]
    \centering
    \includegraphics[width=\columnwidth]{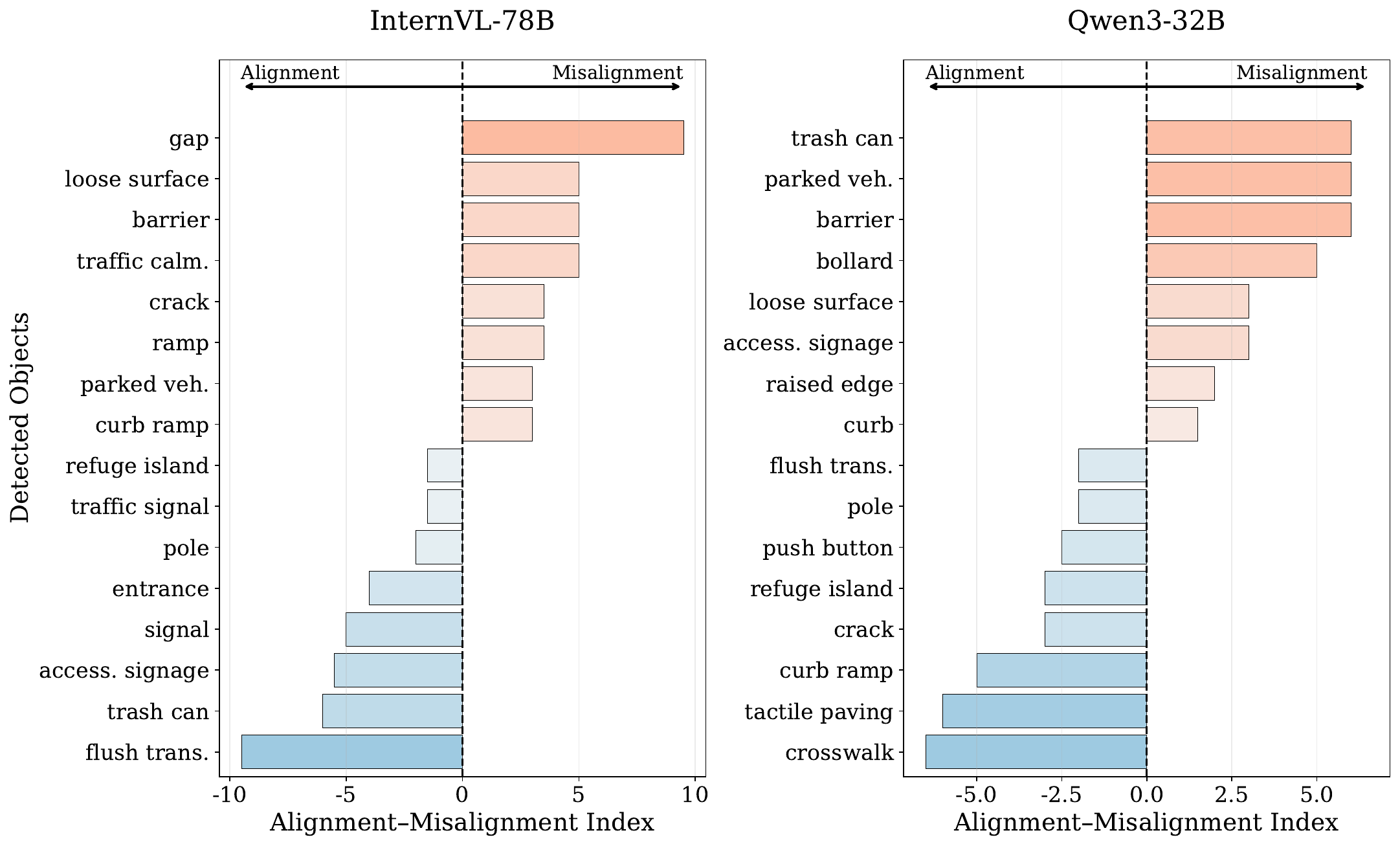}
    \caption{Detected object AMI between well-aligned and poorly-aligned VLM accessibility scores. Positive AMI values indicate semantic objects disproportionately associated with poorly aligned locations (misalignment), whereas negative values indicate stronger association with well-aligned locations (alignment).}
    \label{fig:AMI_alignment}
    \vspace{-4mm}
\end{figure}

\begin{figure}[h]
    \centering
    \includegraphics[width=\columnwidth]{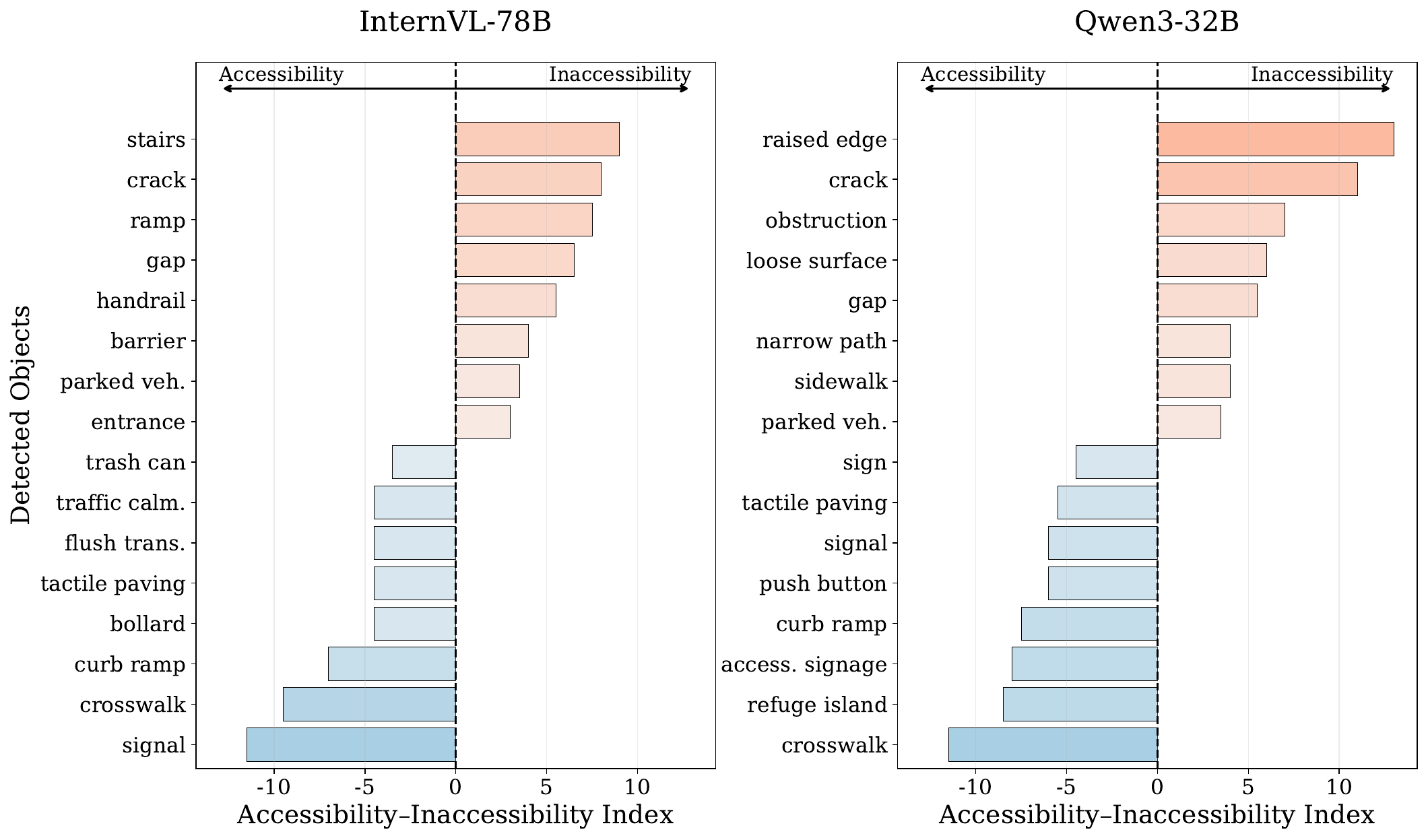}
    \caption{Detected object AII between locations receiving high and low VLM accessibility scores. Positive values indicate semantic objects disproportionately associated with lower VLM accessibility scores, whereas negative values indicate stronger association with higher VLM accessibility scores.}
    \label{fig:AII_scoring}
    \vspace{-4mm}
\end{figure}

\vspace{3px}\noindent\textbf{Visual Cue Association Analysis.} Figures~\ref{fig:AMI_alignment} and \ref{fig:AII_scoring} present complementary rank-difference analyses for VLM--sensor alignment by AMI and accessibility scoring patterns by AII. Across both models, infrastructure features that support physical accessibility such as \textit{crosswalks}, \textit{curb ramps}, \textit{tactile paving}, and \textit{accessible signage} are associated with stronger alignment and higher accessibility scores, whereas degraded pedestrian conditions including \textit{cracks}, \textit{gaps}, \textit{loose surfaces}, \textit{stairs}, \textit{parked vehicles}, and \textit{obstructions} are associated with lower accessibility scores and greater misalignment. Interestingly, InternVL-78B demonstrates greater sensitivity to structural discontinuities, while Qwen3-32B emphasizes surface-level obstacles and pathway constraints. %Overall, the analyses suggest that VLMs capture salient accessibility-supportive infrastructure while remaining challenged by fragmented and visually ambiguous pedestrian environments

%\vspace{3px}\noindent\textbf{Visual Cues: Alignment versus Misalignment.} 
%\VW{VW: Please double-check the reasoning and supporting observations for all visual cues. from DW} 
%Figure~\ref{fig:AMI_alignment} presents the proposed AMI analysis and reveals several consistent semantic patterns across both models. Accessibility-supportive infrastructure such as \textit{crosswalks}, \textit{tactile paving}, and \textit{flush transitions} is more strongly associated with aligned predictions, whereas obstacles and degraded pedestrian conditions including \textit{barriers}, \textit{loose surfaces}, \textit{parked vehicles}, and \textit{cracks} are disproportionately associated with misaligned predictions. InternVL-78B demonstrates greater sensitivity to environmental discontinuities such as \textit{gaps} and \textit{loose surfaces}, while Qwen3-32B places stronger emphasis on street obstacles and urban furniture including \textit{trash cans} and \textit{bollards}. Overall, the results suggest that structured pedestrian guidance features facilitate more reliable multimodal accessibility reasoning, whereas fragmented, obstructed, or degraded sidewalk environments remain challenging for both VLMs.

\section{Discussion and Implications}
\label{sec:discussion}

Our results show that expert-guided VLMs can support scalable accessibility
assessment from Street View imagery, with VLM-derived scores showing
consistent alignment with GPS-derived wheelchair mobility observations. RAG improves
alignment with observed mobility behavior, while the retrieval analysis
shows that VLM scores can help prioritize high-friction locations for
subsequent human review. Visual-cue analysis further indicates stronger
alignment for salient accessibility features, while subtle surface
conditions, transient obstructions, and viewpoint-dependent barriers
remain challenging.

Several limitations should be considered. The study is conducted on a
single campus, and Street View imagery may not reflect current conditions.
Sensor observations are also limited to traversable routes and therefore
may under-represent locations that prevent traversal entirely. In addition,
the mobility data were collected using an able-bodied participant in a
manual wheelchair rather than an actual wheelchair user. Dwell time is
therefore treated as a behavioral proxy for mobility friction rather than
a direct accessibility label. Prolonged dwell may arise from traffic signals, congestion, or navigation decisions, and the study does not independently verify physical accessibility barriers at the evaluated locations. Accordingly, the reported correlations and retrieval performance support preliminary screening rather than definitive barrier identification.

Overall, our findings support the potential of expert-guided VLMs as a
scalable screening tool for prioritizing locations for further accessibility
review. Future work will evaluate the framework across more diverse
environments and incorporate broader input from wheelchair users and
accessibility professionals.
{
    \small
    \bibliographystyle{ieeenat_fullname}
    \bibliography{main}
}

\end{document}